\documentclass[11pt,a4paper]{article}
\usepackage[hyperref]{eacl2021}
\usepackage{times}
\usepackage{latexsym}

\usepackage{microtype}

\aclfinalcopy 

\usepackage{graphicx}
\usepackage{tabularx}
\usepackage{wrapfig}
\usepackage[normalem]{ulem}
\usepackage{amsmath}
\usepackage{subcaption}
\usepackage{xcolor}
\usepackage{multirow}
\usepackage{multicol}
\usepackage{makecell}
\usepackage{comment}
\usepackage{amssymb}
\def\squiggly{\bgroup \markoverwith{\textcolor{red}{\lower3.5\p@\hbox{\sixly \char58}}}\ULon}

\usepackage{url}
\usepackage{xcolor} 
\usepackage{paralist}

\makeatletter
\newcommand{\thickhline}{%
    \noalign {\ifnum 0=`}\fi \hrule height 1pt
    \futurelet \reserved@a \@xhline
}
\makeatother

\title{{Informative} and Controllable Opinion Summarization}

\author{
  Reinald Kim Amplayo \and
  Mirella Lapata \\
  Institute for Language, Cognition and Computation \\
  School of Informatics, University of Edinburgh \\
  reinald.kim@ed.ac.uk, mlap@inf.ed.ac.uk
}

\date{}

\begin{document}
\maketitle

\begin{abstract}
  Opinion summarization is the task of automatically generating
  summaries for a set of reviews about a specific target (e.g.,~a
  movie or a product).  Since the number of reviews for each target can be
  prohibitively large, neural network-based methods follow a two-stage approach where an
  \textit{extractive} step first pre-selects a subset of salient
  opinions and an \textit{abstractive} step creates the summary while
  conditioning on the extracted subset.
  However, the extractive model
  leads to loss of information which may be useful depending on user
  needs.  In this paper we propose a summarization framework that
  eliminates the need to rely only on pre-selected content and waste possibly useful
  information, especially when customizing summaries.  The framework
  enables the use of all input reviews by first \textit{condensing}
  them into multiple dense vectors which serve as input to an
  abstractive model.  We showcase an effective instantiation
  of our framework which produces more informative summaries and also
  allows to take user preferences into account using our zero-shot
  customization technique. Experimental results demonstrate that our model
  improves the state of the art on the Rotten Tomatoes dataset and generates customized summaries effectively.
\end{abstract}

\section{Introduction}
\label{sec:intro}

\begin{figure}[t]
\small
\begin{center}
\begin{tabular}{@{~}p{7.5cm}@{~}} \thickhline
\multicolumn{1}{c}{``Coach Carter'' Reviews}\\\thickhline
\begin{minipage}[t]{7.5cm}
\vspace*{-.2cm}
\begin{asparadesc}
\item[$\bullet$\scshape\bfseries] {\color{green!45!blue}Samuel L. Jackson} plays the real-life coach of a {\color{red!45!blue}high school basketball team} in this solid sports drama ...

\item[$\bullet$\scshape\bfseries]  {\color{green!45!blue}Great performance by Samuel Jackson} but 
 predictable as a slam dunk ...

\item[$\bullet$\scshape\bfseries]  ... excellent 
 {\color{red!45!blue}basketball choreography}, Coach Carter is fun, 
 hopeful, occasionally silly and, what can I say, 
 inspiring.

  \vspace*{-.3cm}
\end{asparadesc}
\end{minipage}\\
\multicolumn{1}{c}{} \\\thickhline
\multicolumn{1}{c}{{Consensus Summary}} \\ \thickhline
Even though it's {based on a true story}, Coach Carter is pretty formulaic stuff,
but it's effective and energetic, thanks to a strong central performance from
Samuel L. Jackson. \\\thickhline
\multicolumn{1}{c}{\textsc{Extract-Abstract} Framework} \\ \thickhline
Coach Carter is a preposterously plotted thriller that \uwave{borrows heavily from other superior films}.
\textit{(factually incorrect)} \\ \thickhline
\multicolumn{1}{c}{\textsc{Condense-Abstract} Framework}\\ \thickhline
\emph{General}: An inspirational flick with a healthy dose of message, but it's too
predictable. \\
\emph{Customized ({\color{green!45!blue}acting})}: An inspirational flick with a 
healthy dose of humor, Coach Carter is a
perceptive sports drama with a {\color{green!45!blue}standout performance
from Samuel L. Jackson.} \\
\emph{Customized ({\color{red!45!blue}plot})}: A feel-good tale with a healthy 
dose of heart, Coach Carter is a worthy
addition to the {\color{red!45!blue}basketball system} that it's difficult to resist. \\\thickhline
\end{tabular}
\end{center}
\caption{Three out of 150 reviews for the movie ``Coach Carter'', and
  summaries written by the editor, and generated by a model following
  the \textsc{Extract-Abstract} approach and the proposed
  \textsc{Condense-Abstract} framework. The latter produces more
  informative and factual summaries whilst allowing to control aspects
  of the generated summary (such as the {\color{green!45!blue}acting}
  or {\color{red!45!blue}plot} of the movie).}
    \label{fig:intro}
\end{figure}

The proliferation of opinions expressed in online reviews, blogs,
and social media has created a pressing need for
automated systems which enable customers and companies
to make informed decisions without having to absorb large
amounts of opinionated text.  Opinion summarization is the task of
automatically generating summaries for a set of opinions about a
specific target \cite{conrad2009query}. Figure~\ref{fig:intro} shows
various reviews about the movie ``Coach Carter'' and example summaries
generated by humans and automatic systems.

The vast majority of previous work \cite{hu2004mining} views opinion
summarization as the final stage of a three-step process involving:
(1)~aspect extraction (i.e.,~finding features pertaining to the target
of interest, such as battery life or sound quality); (2)~sentiment
prediction (i.e., determining the sentiment of the extracted aspects);
and (3)~summary generation (i.e.,~presenting the identified opinions
to the user).  Textual summaries are created following mostly
extractive methods which select representative segments (usually
sentences) from the source text
\cite{popescu2005extracting,blairgoldensohn2008building,lerman2009sentiment}. 
Despite
being less popular, abstractive approaches seem more appropriate for
the task at hand as they attempt to generate summaries which are
maximally informative and minimally redundant without simply
rearranging passages from the original opinions
\cite{ganesan2010opinosis,carenini2013multi,gerani2014abstractive}.

General-purpose summarization approaches have recently shown promising
results with end-to-end models which are data-driven and take
advantage of the success of sequence-to-sequence neural network
architectures.  Most approaches
\cite{rush2015neural,see2017get}
encode documents and then decode the learned representations into an
abstractive summary, often by attending to the source input
\cite{bahdanau2014neural} and copying words from it
\cite{vinyals2015pointer}. Under this modeling paradigm, it is no longer
necessary to identify aspects and their sentiment for the opinion
summarization task, as these are learned \emph{indirectly} from
training data (i.e.,~sets of opinions and their corresponding
summaries). These models are usually tested on domains where the input
is either one document or a small set of documents.

\begin{figure}[t]
    \centering
    \begin{subfigure}{\columnwidth}
        \centering
        \includegraphics[width=\textwidth]{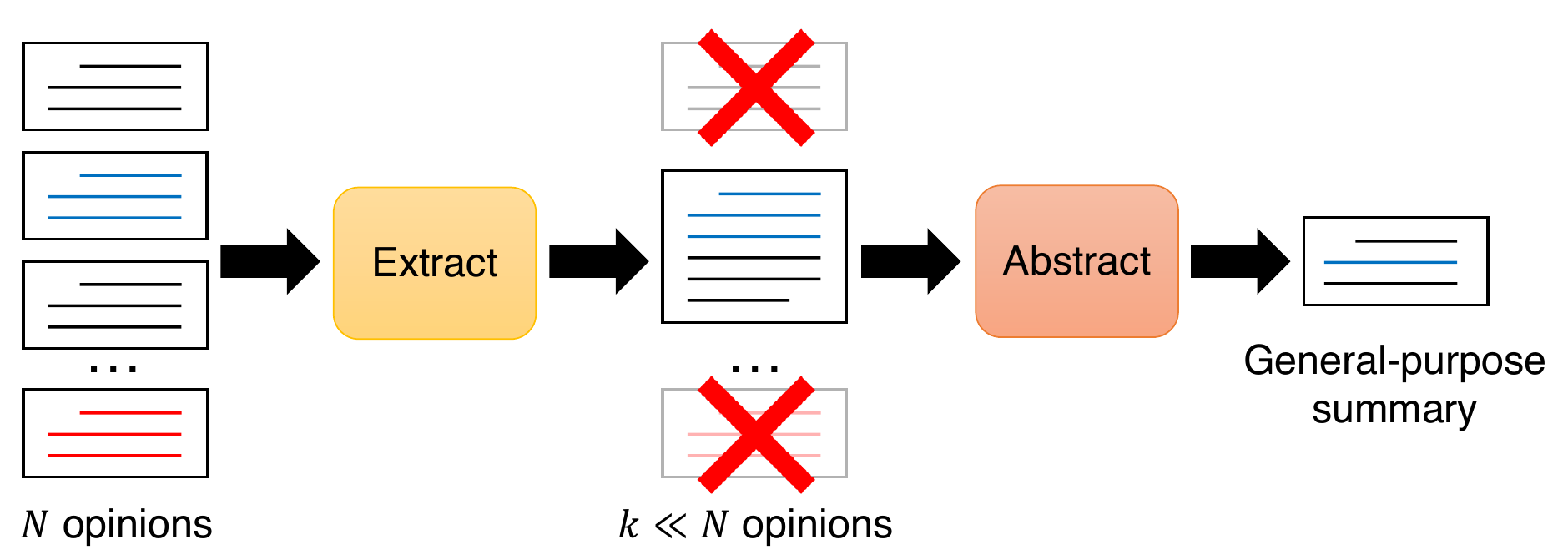}
        \caption{\textsc{Extract}-\textsc{Abstract} (EA) Framework}
        \label{fig:ea_framework}
    \end{subfigure}
    \begin{subfigure}{\columnwidth}
        \includegraphics[width=\textwidth]{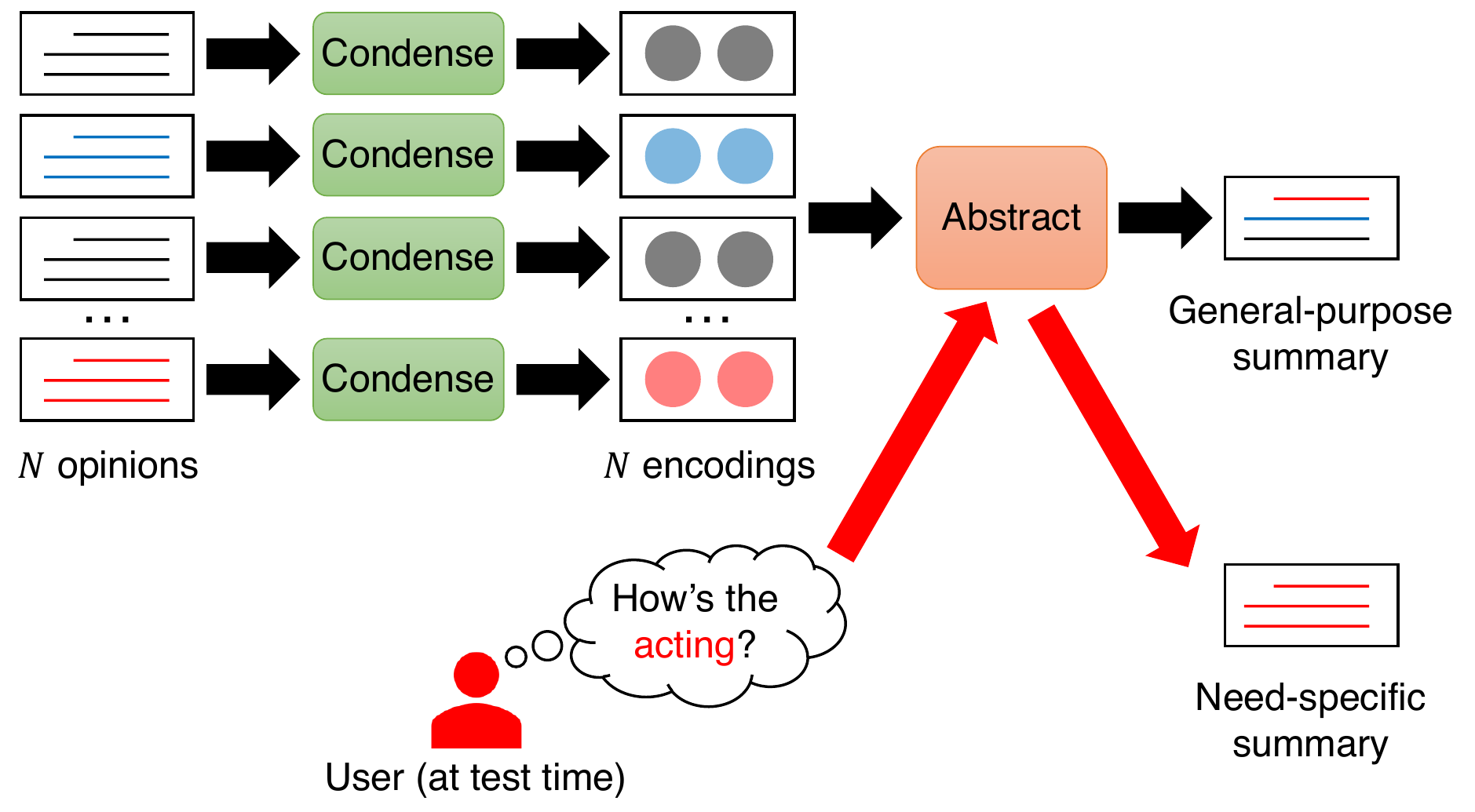}
        \caption{\textsc{Condense}-\textsc{Abstract} (CA) Framework}
        \label{fig:pg_framework}
    \end{subfigure}
    \caption{Illustration of EA and CA frameworks for opinion
      summarization. In the CA framework, users can obtain
      need-specific summaries at test time (e.g.,~give me a summary
      focusing on acting).}
\end{figure}

However, the number of input reviews for each target entity tends to be very large (150 for the
example in Figure~\ref{fig:intro}).  It is therefore practically
unfeasible to train a model in an end-to-end fashion, given the memory
limitations of modern hardware.  As a result, current approaches
\cite{wang2016neural,liu2018generating,liu2019hierarchical}
sacrifice end-to-end elegance in favor of a two-stage framework which
we call \textsc{Extract-Abstract} (EA): an \emph{extractive} model first
selects a subset of opinions and an \emph{abstractive} model then
generates the summary while conditioning on the extracted subset (see
Figure~\ref{fig:ea_framework}). The extractive pass unfortunately has
two drawbacks. Firstly, on account of having access to only a small subset of
reviews, the summaries can be less informative and inaccurate, as
shown in Figure~\ref{fig:intro}.  And secondly, user preferences
cannot be easily taken into account (e.g.,~a user may wish to
obtain a summary focusing on the acting or plot of a movie as opposed
to a general-purpose summary) since more specialized information might
have been removed.

In this paper, we propose \textsc{Condense}-\textsc{Abstract} (CA), an
alternative two-stage framework which enables the use of \emph{all}
input reviews when generating the summary (see
Figure~\ref{fig:pg_framework}).  The \textsc{Condense} model first
represents the input reviews as encodings, aiming to condense their
meaning and distill information relating to sentiment and various
aspects of the target being reviewed.  The \textsc{Abstract} model
then fuses these condensed representations into one aggregate encoding
and generates an opinion summary from it.  We implement a simple yet
effective instantiation of the CA framework, using a vanilla
autoencoder as the \textsc{Condense} model, and a decoder
with attention and copy mechanisms as the \textsc{Abstract} model.  We
also introduce a zero-shot customization technique allowing users to
control important aspects of the generated summary at test time. Our
approach enables controllable generation while leveraging the full
spectrum of opinions available for a specific target.

We perform experiments on a dataset consisting of movie reviews and
opinion summaries elicited from the Rotten Tomatoes website
(\citealp{wang2016neural}; see Figure~\ref{fig:intro}).  Our proposed approach
outperforms state-of-the-art models by a large margin using automatic
metrics and in a judgment elicitation study.  We also verify that our
zero-shot customization technique can effectively generate
need-specific summaries.

\section{Related Work}

Most opinion summarization models follow extractive methods (see
\citealp{kim2011comprehensive} and \citealp{angelidis2018summarizing}
for overviews), with the exception of a few systems which are able to
generate novel words and phrases not featured in the source
text. \citet{ganesan2010opinosis} propose a graph-based framework for
generating concise opinion summaries, while
\citet{gerani2014abstractive} represent reviews as discourse trees
which they aggregate to a global graph to generate a
summary. Other work 
\cite{carenini2013multi,mukherjee2013sentiment}
takes the distribution of opinions and their aspects into account so
as to generate more readable summaries. \citet{difabbrizio2014hybrid}
present a hybrid system which uses extractive techniques to select
salient quotes from the input reviews and embeds them into an
abstractive summary to provide evidence for positive or negative
opinions.

More recent work has seen the effective application of
sequence-to-sequence models
\cite{sutskever2014sequence,bahdanau2014neural} to various abstractive
summarization tasks including headline generation
\cite{rush2015neural}, single-
\cite{see2017get,nallapati2016abstractive}, and multi-document
summarization
\cite{wang2016neural,liu2018generating,liu2019hierarchical}. Closest
to our approach is the work of \citet{wang2016neural} who generate
opinion summaries following a two-stage process which first selects/extracts
reviews bearing pertinent information, and then generates the
summary by conditioning on these reviews. 
More recent models \cite{chu2019meansum,bravzinskas2019unsupervised,amplayo2020unsupervised} perform opinion summarization in an unsupervised way.
However, these are mostly done on toy datasets \cite{chu2019meansum}, typically with a small number of
reviews per target entity.


Our proposed framework 
works better on real-world datasets with a large number of reviews, since it
eliminates the need to rely only on pre-selected salient
reviews which we argue leads to information loss and subsequently less customizable
generation. Instead,
our model first \textit{condenses} the source reviews into
multiple dense vectors which serve as input to a decoder to generate
an abstractive summary. Beyond producing more informative summaries,
we demonstrate that our approach also allows to customize them.  Recent
conditional generation models have focused on controlling various
aspects of the output such as politeness
\cite{sennrich2016controlling}, 
length \cite{kikuchi2016controlling}, 
content
\cite{fan2018controllable}, or style \cite{ficler2017controlling}. In
contrast, our zero-shot customization technique requires
neither training examples of documents and corresponding (customized)
summaries nor specialized pre-processing to encode which tokens in the
input might give rise to customization.



\section{\textsc{Condense}-\textsc{Abstract} Framework}
\label{sec:fram}

We propose an alternative to the \textsc{Extract-Abstract} (EA) approach 
which enables the use of all input reviews when
generating the summary.  Figure~\ref{fig:pg_framework} illustrates our
proposed \textsc{Condense-Abstract} (CA) framework. In lieu of an integrated
encoder-decoder, we generate summaries using two separate models.  The
\textsc{Condense} model returns review encodings for~$N$ input
reviews, while the \textsc{Abstract} model uses these encodings to
create an abstractive summary.  This two-step approach has
two advantages for multi-document summarization. Firstly,
CA-based~models are more
space-efficient, since the set of $N$~reviews is not treated as
one large instance but as $N$~separate instances when training
the \textsc{Condense} model. And secondly, it is possible to generate
maximally informative and customizable summaries targeting specific 
aspects of the input since the
\textsc{Abstract} model operates over the encodings of \emph{all}
available reviews.

In the following subsections, we explain how we instantiate a model using the CA
framework, which we call \textsc{CondaSum}, with an LSTM-based\footnote{
We use LSTMs as our text encoder instead of other popular alternatives, 
such as Transformers \cite{vaswani2017attention}, since LSTMs work better on autoencoder
architectures, as shown in 
the literature \cite{liu2019summae,zhang2020unsupervised}, as well as during our
preliminary experiments.
} vanilla autoencoder
 (\textsc{Condense} model) and a decoder with attention and copy mechanisms
(\textsc{Abstract} model).

\subsection{The \textsc{Condense} Model}

Let $\mathcal{D}$ denote a cluster of $N$ reviews about a specific
target (e.g.,~a movie or product). For each review
$X=\{w_1,w_2,...,w_M\} \in \mathcal{D}$, the \textsc{Condense} model
learns an encoding~$d$, and word-level encodings $h_1, h_2, ..., h_M$.
We employ a Bidirectional Long Short Term Memory (BiLSTM) encoder
\cite{hochreiter1997long} as our \textsc{Condense} model:
\begin{align}
    \label{eq:bilstm_begin}
    \{\overrightarrow{h}_i, \overleftarrow{h}_i\} &= \text{BiLSTM}_f(w_i) \\
    \label{eq:bilstm_end}
    h_i = [\overrightarrow{h}_i;\overleftarrow{h}_i] &\quad 
    d = [\overrightarrow{h}_M;\overleftarrow{h}_1]
\end{align}
where $\overrightarrow{h}_i$ and $\overleftarrow{h}_i$ are forward and
backward hidden states of the BiLSTM at timestep $i$, and~$;$ denotes
concatenation.
  
Training is performed with a reconstruction objective.  We use a
separate LSTM as the decoder where the first hidden state~$z_0$ is set
to~$d$. Words~$w'_t$ are generated using a softmax classifier:
\begin{align}
  z_t &= \text{LSTM}_d(w'_{t-1}, z_{t-1}) \\ \label{eq:hidden-state}
  p(w'_t) &= \text{softmax}(W z_t + b) 
\end{align}
The auto-encoder is trained with a maximum likelihood loss:
\begin{align}
    \mathcal{L}_{condense} &= - \sum\nolimits_{t=1}^M \log p(w_t)
\end{align}

Once training has taken place, we use the
\textsc{Condense} model to obtain $N$~pairs of review
encodings~$\{d_i\}$ and word-level encodings~$\{h_{i,1}, h_{i,2}, ...,
h_{i,M}\}$, $1 \leq i \leq N$ as representations for the reviews in
$\mathcal{D}$.

\subsection{The \textsc{Abstract} Model}
\label{sec:generate}

The \textsc{Abstract} model first fuses the multiple encodings
obtained from the \textsc{Condense} stage and then generates a summary
using a decoder.

\paragraph{Multi-source Fusion}

We aggregate $N$~pairs of review encodings~$\{d_i\}$ and word-level
encodings~$\{h_{i,1}, h_{i,2}, ..., h_{i,M}\}$, $1 \leq i \leq N$ into
a single pair of review encoding~$d'$ and word-level
encodings~$h'_1, h'_2, ..., h'_V$, where~$V$ is the number of total
unique tokens in the input.

Review encodings are fused using an attentive pooling method which
gives more weight to important reviews. Specifically, we learn a set
of weight vectors $a_i \in \mathbb{R}^{D_d}$, where $D_d$ is the
dimension of $d_i$, to weight-sum the review encodings:
\begin{align}
    \label{eq:mean_encoding}
    \bar{d} &= \sum\nolimits_i{d_i} / N \\
    \label{eq:reduce_weights}
    a_i &= \text{softmax}(d_i^\top W_p \bar{d}) \\
    d' &= \sum\nolimits_i a_i * d_i
\end{align}
where the mean encoding $\bar{d}$ is used as the query vector, and
$W_p \in \mathbb{R}^{D_d \times D_d \times D_d}$ is a learned tensor.

We also fuse word-level encodings, since the same words may appear in
multiple reviews. To do this, we simply average all encodings of the
same word, if multiple tokens of the word exist:
\begin{equation}
    h'_j = \sum\nolimits_{(i,k):w_{i,k}=w_j} h_{i,k} / V_{w_j}
\end{equation}
where $V_{w_j}$ is the number of tokens for word $w_j$ in the input.

\paragraph{Decoder}

The decoder generates summaries conditioned on the fused review
encoding $d'$ and word-level encodings
$h'_1,h'_2,...,h'_V$. We use a simple LSTM decoder enhanced with
attention \cite{bahdanau2014neural} and copy mechanisms
\cite{vinyals2015pointer}. We set the first hidden state~$s_0$
to~$d'$, and run an LSTM to calculate the current hidden state using
the previous hidden state~$s_{t-1}$ and word~$y'_{t-1}$ at time
step~$t$:
\begin{equation}
    \label{eq:decoder}
    s_t = \text{LSTM}(y'_{t-1}, s_{t-1})
\end{equation}
At each time step~$t$, we use an attention mechanism over word-level
encodings to output the attention weight vector~$a_t$ and context
vector~$c_t$:
\begin{align}
    \label{eq:att_begin}
    e^i_t &= v^\top \text{tanh}(W_h h'_i + W_s s_t + b_a) \\
    a_t &= \text{softmax} (e_t) \\
    c_t &= \sum\nolimits_i a^i_t * h'_t
    \label{eq:att_end}
\end{align}
Finally, we employ a copy mechanism over the input words to output the
final word probability~$p(y'_t)$ as a weighted sum over the generation
probability~$p_g(y'_t)$ and the copy probability~$p_c(y'_t)$:
\begin{align}
    p_g(y'_t) &= \text{softmax} (W_g [s_t; c_t] + b_g) \\
    \label{eq:copy_begin}
    \sigma_t &= \sigma(v_s^\top s_t + v_c^\top c_t + v_y^\top y'_t) \\
    p_c(y'_t) &= \sum\nolimits_{i:y'_i=y'_t} a^i_t \\
    \label{eq:copy_end}
    p(y'_t) &= 
    \begin{aligned}[t]
        &\sigma_t * p_g(y'_t) +  (1 - \sigma_t) * p_c(y'_t)
    \end{aligned}
\end{align}
where~$W$, $v$, and~$b$  are learned parameters, and $t$~is
the current timestep.

\paragraph{Salience-biased Extracts}

The model presented so far has no explicit mechanism to encourage salience among reviews.  We direct the
decoder towards salient reviews by incorporating information from an
extractive step.  Specifically, we use \textsc{BertCent}, a centroid-based \cite{radev2000centroid} document extraction method 
that obtains document representations by resorting to BERT \cite{devlin2018bert}.

\textsc{BertCent} can be simply described as follows.
Firstly, given a review, we obtain its encoding 
as the average of its token encodings obtained from BERT. 
We then take the average of the review encodings and treat it
as the \textit{centroid} of the input reviews, which approximately represents the information that is considered salient.
We select the top $k$ reviews whose encodings are the
nearest neigbors to the centroid.
The selected reviews are concatenated into a long sequence and encoded using a separate BiLSTM whose output serves as input to
an LSTM decoder.
This decoder generates
a \textit{salience-biased} hidden state~$r_t$. 
We then update hidden state~$s_t$ in
Equation~\eqref{eq:decoder} as $s_t = [s_t; r_t]$.

Using these extracts, we still take all input reviews into account,
while acknowledging that some might be more descriptive than others.
This module is a key component to generating \textit{general-purpose}
opinion summaries, where a set of aspects is deemed more salient than
others (e.g., in general, people care more about the plot rather than
the special effects of a movie).  However, this extractive module may hurt the
customizability of the model (e.g., generating \textit{need-specific} summaries, details explained in Section \ref{sec:custom_module}), which we show in our experiments in
Section~\ref{sec:custom_exp}.

\paragraph{Training}

We use two objective functions to train the \textsc{Abstract} model. Firstly, we use a
maximum likelihood loss to optimize the generation probability
distribution~$p(y'_t)$ based on gold summaries $Y=\{y_1,y_2,...,y_L\}$
provided at training time:
\begin{equation}
    \mathcal{L}_{generate} = - \sum\nolimits_{t=1}^L \log p(y_t)
\end{equation}

Secondly, we propose a way to introduce supervision and guide the
attention pooling weights~$W_p$ in Equation~\eqref{eq:reduce_weights}
when fusing the review encodings. Our motivation is that the
resulting fused encoding~$d'$ should be roughly equivalent to the
encoding of summary~$y$, which can be calculated as
$z=\text{\textsc{Condense}}(y)$. Specifically, we use a hinge loss
that maximizes the inner product between~$d'$ and~$z$ and
simultaneously minimizes the inner product between~$d'$ and~$n_i$,
where~$n_i$ is the encoding of one of five randomly sampled negative
summaries:
\begin{equation}
    \mathcal{L}_{fuse} = \sum\nolimits_{i=1}^5 \text{max}(0, 1 - d' z + d' n_i)
\end{equation}
The final objective is then the sum of both loss
functions:
\begin{equation}
\mathcal{L}_{abstract} = \mathcal{L}_{generate} + \mathcal{L}_{fuse}
\end{equation}

\subsection{Zero-shot Customization}
\label{sec:custom_module}

At test time, we can either generate a general-purpose summary or a
\textit{need-specific} summary. To generate the former, we run the
trained model as is and use beam search to find the sequence of words
with the highest cumulative probability.  To generate the latter, we
employ the following simple technique that revises the query vector~$\bar{d}$ in
Equation~\eqref{eq:mean_encoding}.

More concretely, in the movie review domain, users
might wish to obtain a summary that focuses on a specific sentiment
(positive or negative) or aspect (e.g., acting, plot, etc.) of a movie.
In a different domain, users might care about the price of a
product, its comfort, and so on.  
Since these summaries are not available at training time,
we undertake such customization
without requiring access to need-specific summaries. 
Instead, at test time, we assume access to background reviews to
represent the user need.  For example, if we wish to generate a
positive summary, our method requires a set of reviews with positive
sentiment. This is an easy and practical way to approximately provide the model some background 
on how sentiment is communicated in a review.

We use these background reviews conveying a user need~$x$
(e.g.,~acting, plot, positive or negative sentiment) in the
multi-source fusion module to attend more to input reviews related
to~$x$.  Let~$C_x$ denote the set of background reviews.  We obtain a
new query vector $\hat{d} = \sum_{c=1}^{|C_x|} d_c / |C_x|$,
where~$d_c$ is the encoding of the $c$'th review in~$C_x$,
calculated using the \textsc{Condense} model. This simple change allows the
model to focus on input reviews with semantics similar to the user's
need as conveyed by the background reviews~$C_x$.  The new query
vector $\hat{d}$ is used instead of $\bar{d}$ to obtain review
encoding~$d'$ (see Equation~\eqref{eq:mean_encoding}).

\section{Experimental Setup}

\paragraph{Dataset} 
We performed experiments on the Rotten Tomatoes
dataset\footnote{\url{http://www.ccs.neu.edu/home/luwang/publications.html}}
provided in \citet{wang2016neural}. It contains 3,731 movies; for each
movie we are given a large set of reviews written by
professional critics and users and a gold-standard consensus summary
written by an editor (see an example in Figure~\ref{fig:intro}).
We report the dataset statistics in Table \ref{tab:dataset}.
Following
previous work \cite{wang2016neural}, we used a generic label for movie
titles during training which we replace with the original titles 
during inference.

\begin{table}[t]
    \centering
    \begin{tabular}{@{}lrrr@{}}\thickhline
          &    Train & Dev &  Test \\      \thickhline
        \#movies & 2,458  & 536 & 737 \\
        \#reviews/movie &  100.0 & 98.0 & 100.3 \\
        \#tokens/review & 23.6 &  23.5 & 23.6 \\
        \#tokens/summary & 23.8 & 23.6 & 23.8 \\
        \thickhline
    \end{tabular}
    \caption{Dataset statistics of Rotten Tomatoes.} 
    \label{tab:dataset}
\end{table}

\begin{table*}[t]
  \centering
    \begin{tabular}{@{}lccccc@{}}
    \thickhline
    \multicolumn{1}{@{}c}{Model} & METEOR & ROUGE-SU4  & ROUGE-1    & ROUGE-2    & ROUGE-L \\
    \thickhline    
 \textsc{LexRank}* & 5.59  & 3.98  & 14.88 & 1.94 & 10.50 \\
  \textsc{Opinosis}* & 6.07  & 4.90  & 14.98 & 3.07 & 12.19 \\
   \textsc{SummaRunner} & {7.44}  & {5.50}  &  {15.86} & {2.55}  & {12.15} \\ 
	\textsc{BertCent}  & 8.89 & 7.13  & 17.65 & 2.78 & 12.78 \\  \hline
    \textsc{Regress+S2S}* & {6.51}  & {5.70}  & ---     & ---     & --- \\
    \textsc{BertCent+S2S} & 7.42 & 6.61 & 17.59 & 7.34 & 15.83 \\
    \textsc{BertCent+PtGen} & 8.15 & 6.99 & 19.71 & 7.43 & 17.25 \\
	\textsc{CondaSum} & \textbf{8.90} & \textbf{7.79} & \textbf{22.49} & \textbf{7.65} & \textbf{18.47} \\
    \thickhline
    \end{tabular}%
    \caption{Automatic evaluation results on models trained on the original training data. 
      Models whose METEOR and ROUGE-SU4 results are
      taken from \citet{wang2016neural} are marked with an
      asterisk~*. Best performing results per metric are \textbf{boldfaced}.}
  \label{tab:auto_results}%
\end{table*}%

\paragraph{Training Configuration}
For all experiments, our model used word embeddings with
128~dimensions, pretrained using GloVe \cite{pennington2014glove}. We
set the dimensions of all hidden vectors to~256 and the batch size to~8.
For decoding summaries, we use a length-normalized beam search with beam size of~5. 
We applied dropout
\cite{srivastava2014dropout} at a rate of~0.5. The model was trained
using the Adam optimizer \cite{kingma2015adam} with default parameters and $l_2$ constraint \cite{hinton2012improving}
of~2. We performed early stopping based on model performance on the
development set. Our model is implemented in PyTorch\footnote{Our code
  can be downloaded from \url{xxx.yyy.zzz}.}.

\paragraph{Comparison Systems} 

We compare our approach against two types of methods: one-pass
methods and methods that use the EA framework.  One-pass methods
include (a) \textsc{LexRank} \cite{erkan2004lexrank}, a PageRank-like
summarization algorithm which generates a summary by selecting the
$n$~most salient units, until the length of the target summary is
reached; (b)~\textsc{Opinosis}
\cite{ganesan2010opinosis}, a graph-based abstractive summarizer that
generates concise summaries of highly redundant opinions; 
(c)~\textsc{SummaRunner} \cite{nallapati2017summarunner}, a supervised neural
extractive model where each review is classified as to whether it
should be part of the summary or not; and (d)~\textsc{BertCent}, a centroid-based method discussed in Section~\ref{sec:generate} that selects $k=1$ review nearest to the centroid.

EA-based methods include (g)~\textsc{Regress+S2S}
\cite{wang2016neural}, an instantiation of the EA framework where a
ridge regression model with hand-engineered features implements the
\textsc{Extract} model, while an attention-based sequence-to-sequence
neural network is the \textsc{Abstract} model;
(h)~\textsc{BertCent+S2S}, our implementation of an EA-based system
which uses \textsc{BertCent} instead of \textsc{Regress} as the
\textsc{Extract} model; and (i)~\textsc{BertCent+PtGen}, the
same model as (h)~but enhanced with a copy mechanism
\cite{vinyals2015pointer}. For all extractive steps, we set $k=5$,
which is tuned on the development set.

\section{Results}
\label{sec:exp}

\paragraph{Automatic Evaluation}

We considered two evaluation metrics which are also reported in
\citet{wang2016neural}: METEOR \cite{denkowski2014meteor}, a
recall-oriented metric that rewards matching stems, synonyms, and
paraphrases, and ROUGE-SU4 \cite{lin2004rouge} which is calculated as
the recall of unigrams and skip-bigrams up to four words. We also
report F$_{1}$-scores for ROUGE-1/2/L \cite{lin2004rouge}.
Unigram and bigram overlap (ROUGE-1
and ROUGE-2) are a proxy for assessing informativenes while the
longest common subsequence (ROUGE-L) measures fluency.

Our results are presented in Table~\ref{tab:auto_results}.  
Among one-pass systems, the extractive model \textsc{BertCent}
performs the best; despite being unsupervised and extractive,
it benefits from the ability of large neural language models to
learn general-purpose representations. When used in EA-based systems, 
\textsc{BertCent} also improves the system performance, where
\textsc{BertCent+PtGen} performs the best.
Interestingly, \textsc{BertCent} performs better than 
\textsc{BertCent+PtGen} in terms of METEOR and ROUGE-SU4, while the latter performs better in terms of ROUGE-1/2/L.
Our CA-based model \textsc{CondaSum} outperforms all other models
across all metrics, showing that exploiting information about
all reviews helps in improving performance.

We present in Table~\ref{tab:ablation} various ablation studies, which assess the contribution of different model components. Results confirm that our multi-source fusion method and the fusion loss improve performance.
Morevoer, using \textsc{BertCent} for the salient-biased extractive step is better than no extractive step or using 
\textsc{SummaRunner}, which is a weaker extractive model.
Both multi-source fusion and salient-biased extracts help create better general-purpose summaries; 
the former learns which reviews to focus on while the latter explicitly selects
the most important ones.

\begin{table}[t]
	\centering
	\begin{tabular}{@{}lc@{}}
	\thickhline
	\multicolumn{1}{c}{Model} & ROUGE-L \\
	\thickhline
	\textsc{CondaSum} & 18.47 \\
	\hline
	\quad Mean document fusion & 16.69 \\
	\quad No fusion loss & 15.10 \\
	\hline
	\quad No salience-biased extracts & 16.44 \\
	\quad \textsc{SummaRunner} extracts & 17.80 \\
	\thickhline
	\end{tabular}
        \caption{ROUGE-L of \textsc{CondaSum} with less effective document fusion method (second block) and without using our salience-biased extractive step (third block). See
          Appendix for more detailed comparisons.}
  \label{tab:ablation}%
\end{table}

\paragraph{Human Evaluation}

\begin{table}[t]
  \centering
    \begin{tabular}{@{}l@{~~~}rr@{~~~}c@{}}
    \thickhline
    Model & \multicolumn{1}{c}{Inf} & \multicolumn{1}{c}{Corr} & \multicolumn{1}{c}{Gram}  \\
    \thickhline
    \textsc{BertCent+PtGen} & -0.263 & -0.358 & \hspace*{.15cm}-0.152${}^*$ \\
    \textsc{BertCent} & -0.179 & -0.112 & \hspace*{.15cm}-0.102${}^*$ \\
     \textsc{CondaSum} & {-0.042} & {0.021} & -0.078 \\
    \textsc{Gold}  & 0.483 & 0.448 & \hspace*{.15cm}0.331 \\    \thickhline
    \end{tabular}%
  \caption{Best-worst scaling scores on informativeness (Inf), correctness (Corr) and grammaticality (Gram). All pairwise systems differences between \textsc{CondaSum} and other system summaries are significant, except the values marked with asterisk (*),
      based on a one-way ANOVA
      with posthoc Tukey HSD tests ($p<0.05$).}
  \label{tab:human_results}%
\end{table}%

In addition to automatic evaluation, we also assessed system output by
eliciting human judgments.  Participants compared summaries produced
from the best extractive baseline (\textsc{BertCent}), the best
EA system (\textsc{BertCent+PtGen}), and
 our model \textsc{CondaSum}, respectively. As an upper bound, we also
included \textsc{Gold} standard summaries.

The study was conducted on the Amazon Mechanical Turk platform using
Best-Worst Scaling (BWS; \citealp{louviere2015best}), a less
labor-intensive alternative to paired comparisons that has been shown
to produce more reliable results than rating scales
\cite{kiritchenko2017best}. Specifically, participants were shown the
movie title and basic background information (i.e.,~synopsis, release
year, genre, director, and cast).  They were also presented with three
system summaries and asked to select the \textit{best} and
\textit{worst} among them according to three criteria: \emph{Informativeness}
(i.e.,~does the summary convey opinions about specific aspects of the
movie in a concise manner?), \emph{Correctness} (i.e., is the
information in the summary factually accurate and corresponding
to the information given about the movie?), and \emph{Grammaticality}
(i.e.,~is the summary fluent and grammatical?). Examples of
summaries are shown in Figure~\ref{fig:intro} and more can be found
in the Appendix. We randomly selected 50 movies from
the test set and compared all possible combinations of summary triples
for each movie. We collected three judgments for each comparison. The
order of summaries and movies was randomized per participant.

The scores are computed as the percentage of times it was
chosen as best minus the percentage of times it was selected as
worst. The scores range from -1 (worst) to 1 (best) and are shown in
Table~\ref{tab:human_results}. Perhaps unsurprisingly, the
human-generated gold summaries were considered best, whereas our model
\textsc{CondaSum} was ranked second, indicating that humans
find its output more informative, correct, and grammatical compared to
other systems. \textsc{BertCent} was ranked third followed by
\textsc{BertCent+PtGen}. We inspected the summaries produced
by the latter system and found they were factually incorrect bearing
little correspondence to the movie (examples shown in the Appendix), 
possibly due to the huge
information loss at the extraction stage.

\paragraph{Customizing Summaries}
\label{sec:custom_exp}

We further assessed the ability of CA systems to generate customized
summaries at test time.  We evaluate \textsc{CondaSum} models with
and without the salience-biased extractive step. 
The latter model biases summary generation
towards the $k$~most salient extracted opinions {using an additional
  extractive module} which may discard information relevant to the
user's need.  We thus expect this model to be less effective for
customization than \textsc{CondaSum} which makes no assumptions
regarding which summaries to consider.

In this experiment, we assume users may wish to control the output
summaries in four ways focusing on acting- and plot-related aspects of
a movie review, as well as its sentiment, which may be positive or
negative.  Let \textsc{Cust}($x$) be the zero-shot customization
technique discussed in the Section~\ref{sec:custom_module}, where $x$
is an information need (i.e.,~acting, plot, positive, or negative).
We sampled a set of background reviews $C_x$ ($|C_x|$=1,000)
from a corpus of 1 million reviews covering 7,500 movies from the
Rotten Tomatoes website, made available in
\citet{ficler2017controlling}.  The reviews contain sentiment labels
provided by their authors and heuristically classified aspect
labels.  We then
ran \textsc{Cust}($x$) using both the \textsc{CondaSum} models.  We show in
Figure~\ref{tab:example_custom} customized summaries generated by the
models.

\begin{figure}[t]
  \small
\begin{tabular}{@{}p{7.7cm}@{}}\thickhline
 \multicolumn{1}{c}{\textbf{ \textsc{Gold}}}\\
Whether you choose to see it as a statement on consumer culture or simply a special effects-heavy popcorn flick, Gremlins is a minor classic. \\
\hline
 \multicolumn{1}{c}{\textbf{ \textsc{CondaSum} with extractive step}}\\
      \textit{{General}}: Gremlins is a wholesome, entertaining horror film with an enormous cast of eager stars. \\
      \textit{{Customized (Positive)}}: Gremlins is a {\color{blue!45!blue}wholesome, entertaining horror film with an enormous cast of eager stars.} \\
      \textit{{Customized (Negative)}}: Gremlins is a {\color{blue!45!blue}wholesome, entertaining horror film with an enormous cast of eager stars.} \\
 \hline
\multicolumn{1}{c}{ \textbf{ \textsc{CondaSum} without extractive step}}\\
\emph{{General}}: Gremlins may appeal to the dark Christmas horror genre. \\
     \emph{{Customized (Positive)}}: Gremlins is {\color{blue!45!blue} an intelligent, funny Christmas horror film} from Joe Dante's novel. \\
     \emph{{Customized (Negative)}}: Gremlins is {\color{red!45!red} an
        atrociously-acted project whose unoriginal and ineptly-staged
        horror film} from Joe Dante's novel.
         \\
         \thickhline
\end{tabular}
  \caption{
 Examples of general-purpose and need-specific opinion
  summaries for the movie ``Gremlins'', generated by two versions of \textsc{CondaSum}. We
  also show the consensus summary (\textsc{Gold}).  Words/phrases in
  color highlight aspects pertaining to {\color{blue!45!blue}
    positive} and {\color{red!45!red} negative}. More examples can
  be found in the Appendix.}
\label{tab:example_custom}
\end{figure}

To determine which system is better at customization, we again
conducted a judgment elicitation study on Amazon Mechanical Turk.  Participants read a
summary which was created by a general-purpose system or its
customized variant. They were then asked to decide if the summary is
generic or focuses on a specific aspect (plot or acting) and expresses
positive, negative, or neutral sentiment.  We selected 50~movies (from
the test set) which had mixed reviews and collected judgments from
three different participants per summary. The summaries were presented
in random order per participant.

\begin{table}[t]                      
\center
\begin{tabular}{@{}lcccc@{}} 
\thickhline
& \multicolumn{2}{c}{with extracts} &
\multicolumn{2}{c@{}}{without extracts} \\
Customized &   {No}    & {Yes} &   {No}    & {Yes} \\  \thickhline
Acting        & {40.3} & {40.3}  & {42.0}  &  \textbf{{78.0}}  \\
Plot          & {73.3} & {75.0}  & {51.3}  &  \textbf{{76.7}}  \\
Positive      & {66.0} & {67.7}  & {65.3}  &  \textbf{{80.0}}  \\ 
Negative      & {22.7} & {22.0} & {20.7}  &   \textbf{{40.7}} \\ \thickhline   
\end{tabular}
\caption{Proportion of summaries which mention a specific
  aspect/sentiment. \textbf{Boldfaced} values show a significant
  increase ($p<0.01$; using two-sample bootstrap tests) compared to
  the non-customized system variant. 
 Aspects are not mutually exclusive (e.g.~a
  summary may talk about both acting and plot), thus the total
  percentage may exceed 100\%.} 
  \label{tab:cust_result}
\end{table}

Table~\ref{tab:cust_result} shows what participants thought of
summaries produced by non-customized systems (see column No) and
systems which had customization switched on (see column Yes). Overall,
we observe that \textsc{CondaSum} without the extractive step is able to customize summaries to a
great extent. In all cases, crowdworkers perceive a significant
increase in the proportion of aspect $x$ when using
\textsc{Cust}($x$).  \textsc{CondaSum} with the extractive step is unable to generate
need-specific summaries, showing no discernible difference between
generic and customized summaries. This indicates that the use of an
extractive module, which is one of the main components of EA-based
approaches, limits the flexibility of the abstractive model to
customize summaries based on a user need.

\section{Conclusions}
\label{sec:conclusions}

We introduced the \textsc{Condense-Abstract} (CA) framework for
opinion summarization which eliminates the need to rely only on a small subset of extracted reviews and allows the use of all reviews to generate maximally informative
summaries.  We presented \textsc{CondaSum}, an instantiation of this
framework and showed in
both automatic and human-based evaluation that it is superior to
purely extractive models and abstractive models that include an
extractive pre-selection stage.  We also showed that when an extractive step is not used, our zero-shot
customization technique is able to generate need-specific summaries at
test time.  
In the future, we plan to apply the CA framework to other
multi-document summarization tasks.

\bibliography{eacl2021.bib}
\bibliographystyle{acl_natbib}


\clearpage
\appendix
\section{Appendices}

\subsection{Ablation Studies}

We performed ablation studies on \textsc{CondaSum} to four different versions: (a) using a mean document fusion instead of our multi-source fusion module, (b) without using a fusion loss, (c) without using salience-biased extracts, and (d) using outputs from \textsc{SummaRunner} \cite{nallapati2017summarunner} as salience-biased extracts. Table~\ref{tab:ablation_full} shows the ROUGE-1/2/L F1-scores of our model and various versions thereof. The final model consistently
performs better on all metrics.

\begin{table*}[t]
  \centering
    \begin{tabular}{@{}lccccc@{}}
    \thickhline
    \multicolumn{1}{c}{Model} & METEOR & ROUGE-SU4 & ROUGE-1 & ROUGE-2 & ROUGE-L \\
    \thickhline
    \textsc{CondaSum} & \textbf{8.90} & \textbf{7.79} & \textbf{22.49} & \textbf{7.65} & \textbf{18.47} \\
    \hline
    \quad Mean document fusion & 8.21 & 6.63 & 19.99 & 6.43 & 16.69 \\
    \quad No fusion loss & 8.09 & 6.23 & 18.57 & 5.12 & 15.10 \\
    \hline
    \quad No salience-biased extracts & 8.56 & 6.81 & 20.22 & 6.17 & 16.44 \\
    \quad \textsc{SummaRunner} extracts & 8.50 & 7.39 & 21.19 & 7.64 & 17.80 \\
\thickhline
    \end{tabular}%
    \caption{ROUGE-1/2/L F1 scores of our model and versions thereof with less effective document fusion method (second block) and without using our salience-biased extractive step (third block).}
  \label{tab:ablation_full}%
\end{table*}%

\subsection{Amazon Mechanical Turk Human Evaluation Experiments}

We conducted three different human evaluation experiments: the
best-worst scaling evaluation, the aspect-specific (acting vs. plot)
customization evaluation, and the sentiment-specific (positive
vs. negative) customization evaluation. To lessen the burden on the
annotators and consequently gather more accurate responses, we
conducted three separate Amazon Mechanical Turk (AMT) experiments. For
all experiments, we ensure turkers have an approval rate of 98\% (or
above) with at least 1,000 tasks approved. Furthermore, turkers should
be a (self-reported) native English speakers from one of the following
countries: Australia, Canada, Ireland, New Zealand, United Kingdom,
and United States.  We discuss specific configurations for each
experiment in the next paragraphs.

\paragraph{Best-Worst Scaling Evaluation}

Each Human Intelligence Task (HIT) consists of five questions a turker
must answer to receive payment. Each question includes the title of
the movie, and corresponding basic background information: synopsis,
release year, genre, director, and actors (see examples in
Figures~\ref{fig:app_ex_aspect} and~\ref{fig:app_ex_sentiment}). The
summaries shown are randomly shuffled and are labeled A, B, or C.
Turkers are then asked which to select the best or worst summary
amongst A, B, and C, according to informativeness (i.e., does the
summary convey opinions about specific aspects of the movie in a
concise manner?), correctness (i.e., is the information in the summary
factually accurate and corresponding to the information given about
the movie?), and grammaticality (i.e., is the summary fluent and
grammatical?).  The criteria and their definitions are shown to
turkers to guide them while they select their answers.

\paragraph{Aspect-Specific Customization Evaluation}

Similar to the best-worst scaling template, each HIT also consists of
five questions a turker must answer to receive payment. Each question
also includes the movie title and its basic background
information. Turkers are given a summary which they have to read. They
then choose the most appropriate answer from the following choices:
(a) mentions neither acting nor plot, (b) mentions acting, (c)
mentions plot, or (d) mentions both acting and plot.

\paragraph{Sentiment-Specific Customization Evaluation}

For this evaluation, the template is similar with that of the
aspect-specific customization evaluation, but the choices are
different. The turkers instead are given three choices: (a) neutral
sentiment, (b) positive sentiment, and (c) negative sentiment.

\subsection{Example Summaries}

Finally, we show more example system summaries generated by
\textsc{SummaRunner}, \textsc{SummaRunner+PtGen}, \textsc{CondaSum},
and \textsc{CondaSum+Salient}, together with the \textsc{Gold} summary
in Figures \ref{fig:app_ex_aspect}--\ref{fig:app_ex_sentiment}. Figure
\ref{fig:app_ex_aspect} additionally shows summaries that are
customized based on the plot or acting aspect of a movie, while Figure
\ref{fig:app_ex_sentiment} shows customized summaries according to
positive or negative sentiment. The examples show similar trends with
the examples in the main paper.

\begin{figure*}[t]
  \centering
\begin{small}
    \begin{tabular}{@{}lp{12.2cm}@{}}
      \thickhline
      \multicolumn{2}{c}{Movie: ``Kitchen Stories''} \\ 
	\thickhline
      Synopsis & Director Bent Hamer's comedy drama Salmer Fra Kjøkkenet (Kitchen Stories) is based on the real-life social experiments conducted in Sweden during the 1950s. In the years following WWII, a research institute sets out to modernize the home kitchen by observing a handful of rural Norwegian bachelors. In the small town of Landstad, middle-aged Isak (Joachim Calmeyer) is one such research subject who regrets ever agreeing to participate in the study. Nevertheless, he is observed by Folke (Tomas Norström), and the two develop a strange friendship until the observer becomes sick. This causes a problem with Folke's boss (Reine Brynolfsson) and Isak's friend Grant (Bjørn Floberg). \\
      Year & 2004 \\
      Genre & Art House \& International, Comedy, Drama \\
      Director & Bent Hamer \\
      Actors & Joachim Calmeyer as Isak Bjorvik, Tomas Norström as Folke Nilsson, Bjørn Floberg as Grant, Reine Brynolfsson as Malmberg, Sverre Anker Ousdal as Dr. Benjaminsen \\      
      \hline
      {\textsc{\textbf{Gold}}} & 
      By turns touching and funny, this Norwegian import offers quietly absorbing commentary on modern life and friendship. \\
      {\textsc{\textbf{BertCent}}} & 
      An extended ethnic joke, Kitchen Stories moves with a glacial indifference to conventional comedic timing. But perhaps that's what makes it so funny and so emotionally precise. \\
      {\textsc{\textbf{BertCent+PtGen}}} & 
      Kitchen Stories is a smart, funny social \uwave{satire about modern-day Jerusalem.} \\
      {\textbf{\textsc{CondaSum} w/o extracts}} & 
      \emph{\textbf{General}}: Kitchen Stories is an {offbeat, thought-provoking tale} that's both funny and moving. \\
&     \emph{\textbf{Customized (Acting)}}: Kitchen Stories is an intelligent, funny social comedy that {\color{green!45!blue} benefits from an impressive cast and outstanding performances from Isak}. \\
 &    \emph{\textbf{Customized (Plot)}}: Kitchen Stories is both funny and
      smart, {\color{red!45!blue} featuring a highly original
        script}. \\
      {\textbf{\textsc{CondaSum} w/ extracts}}  & 
      \textit{\textbf{General}}: 
        Kitchen Stories is a well-acted, offbeat comedy with a fine performance from Isak Bjorvik. \\
     & \textit{\textbf{Customized (Acting)}}: 
        Kitchen Stories is a well-acted, offbeat comedy with {\color{green!45!blue} a fine performance from Isak Bjorvik}.\\
 &     \textit{\textbf{Customized (Plot)}}: 
         Kitchen Stories is a well-acted, offbeat comedy with {\color{green!45!blue} a fine performance from Isak Bjorvik}.\\
    \thickhline 
      \multicolumn{2}{c}{Movie: ``Get Smart''} \\ 
	\thickhline
      Synopsis & 40-Year-Old Virgin star Steve Carell steps into the telephonic shoes of television's most beloved bumbling detective in this big-screen adaptation of the hit 1960s-era comedy series created by Mel Brooks. The evil geniuses at KAOS have hatched a diabolical plot to dominate every living man, woman, and child on the planet, and their plot gets under way as they attack the headquarters of the U.S. spy agency Control. As a result of the attack, the identity of every agent working for Control has been compromised. Realizing that the only way to thwart KAOS' evil plan is to promote eager but inexperienced Control analyst Maxwell Smart (Carell) to the rank of special agent, the Chief (Alan Arkin) reluctantly teams Smart with Agent 99 (Anne Hathaway) -- a veteran super-spy whose beauty is only surpassed by her lethality. With no real field experience to speak of and nothing but sheer enthusiasm and a handful of fancy spy gadgets to help him accomplish his deadly mission, Maxwell Smart his new partner, Agent 99, will be forced to faces malevolent KAOS head Siegfried (Terence Stamp) and his loyal army of minions in a decisive fight that will determine the fate of the free world. Dwayne "The Rock" Johnson, David Koechner, Terry Crews, and Ken Davitian co-star. \\
      Year & 2008 \\
      Genre & Action \& Adventure, Comedy \\
      Director & Peter Segal \\
      Actors & Steve Carell as Maxwell Smart, Anne Hathaway as Agent 99, Alan Arkin as The Chief, Terence Stamp as Siegfried, Terry Crews as Agent 91\\     
      \hline
      {\textsc{\textbf{Gold}}} & 
      Get Smart rides Steve Carell's considerable charm for a few laughs, but in the end is a rather ordinary summer comedy. \\
      {\textsc{\textbf{BertCent}}} & 
      Although Carell is never less than likable, he's funnier in any random scene of the office. Here's hoping some misguided team doesn't try to turn that series into a quick grab at box-office bucks 40 years from now. \\
      {\textsc{\textbf{BertCent+PtGen}}} & 
      Get Smart is a smart, funny, funny, and entertaining \uwave{legal thriller}.\\
      {\textbf{\textsc{CondaSum} w/o extracts}} & 
      \emph{\textbf{General}}: Get Smart is an atrociously-acted project which is unoriginal. \\
&     \emph{\textbf{Customized (Acting)}}: Get Smart is an atrociously-acted project which is unoriginal, but ineptly-staged action sequences remind viewers of {\color{green!45!blue} Steve Carell's knockout performance}. \\
 &    \emph{\textbf{Customized (Plot)}}: Get Smart is a preposterously inept  \uwave{psychological thriller} that {\color{red!45!blue} borrows heavily from other superior films}. \\
      {\textbf{\textsc{CondaSum} w/ extracts}}  & 
      \textit{\textbf{General}}: 
      Get Smart is a silly, breezy, and funny comedy that will reward patient viewers. \\
     & \textit{\textbf{Customized (Acting)}}: Get Smart is a silly, breezy, and funny {\color{red!45!blue} comedy that will reward patient viewers}. \\
 &     \textit{\textbf{Customized (Plot)}}: Get Smart is a silly, breezy, and funny {\color{red!45!blue} comedy that will reward patient viewers}. \\ \thickhline
    \end{tabular}%
\end{small}
\caption{Examples of general-purpose and need-specific opinion summaries
  generated by four systems. We also show the consensus summary
  (\textsc{Gold}).  \uwave{Underlined} phrases denote factually
  incorrect information.  Words/phrases in color highlight aspects
  pertaining to {\color{green!45!blue}
    acting} and {\color{red!45!blue} plot}.}
  \label{fig:app_ex_aspect}%
\end{figure*}%

\begin{figure*}[t]
  \centering
\begin{small}
    \begin{tabular}{@{}lp{12.2cm}@{}}
      \thickhline
      \multicolumn{2}{c}{Movie: ``Ladder 49''} \\ 
	\thickhline
      Synopsis & Baltimore firefighter Jack Morrison, making the transition from inexperienced rookie to seasoned veteran, struggles to cope with a risky, demanding job that often shortchanges his wife and kids. He relies on the support of his mentor and captain, Mike Kennedy and his second family--found in the brotherly bond between the men of the firehouse. But when Jack becomes trapped in the worst blaze of his career, his life and the things he holds important--family, dignity, courage--come into focus. As his fellow firemen of Ladder 49 do all they can to rescue him, Jack's life hangs in the balance. \\
      Year & 2004 \\
      Genre & Action \& Adventure, Drama \\
      Director & Jay Russell \\
      Actors & Joaquin Phoenix as Jack Morrison, John Travolta as Capt. Mike Kennedy, Jacinda Barrett as Linda Morrison, Robert Patrick as Lenny Richter, Morris Chestnut as Tommy Drake\\     
      \hline
      {\textsc{\textbf{Gold}}} & 
      Instead of humanizing the firemen, the movie idolizes them, and thus renders them into cardboard characters. \\
      {\textsc{\textbf{BertCent}}} & 
      It piles on the schmaltz and the lame attempts at humor so that we'll see the film as a comedy with a big heart, a really big heart. \\
      {\textsc{\textbf{BertCent+PtGen}}} & 
      Ladder 49 is a lightweight, formulaic \uwave{tween version of fame}.\\
      {\textbf{\textsc{CondaSum} w/o extracts}} & 
      \emph{\textbf{General}}: Phoenix's Jack Morrison performance makes Ladder 49 something like a rote old-fashioned adventure. \\
&     \emph{\textbf{Customized (Positive)}}: Phoenix's Jack Morrison performance makes Ladder 49 something like a {\color{red!45!red} rote old-fashioned adventure flick}, but {\color{blue!45!blue}with a juicy story}. \\
 &    \emph{\textbf{Customized (Negative)}}: Phoenix's Jack Morrison performance makes Ladder 49 something like a {\color{red!45!red} rote old-fashioned adventure flick, and it's ultimately too rough and uneven to hang together} as a wholly satisfying viewing experience. \\
      {\textbf{\textsc{CondaSum} w/ extracts}}  & 
      \textit{\textbf{General}}: 
      Ladder 49 is a smart, tender, and poignant drama about hopeless love. \\
     & \textit{\textbf{Customized (Positive)}}: 
     Ladder 49 is a {\color{blue!45!blue}sentimental and sincere tale} that resonates with truth.\\
 &     \textit{\textbf{Customized (Negative)}}: 
    Ladder 49 is a {\color{blue!45!blue}sentimental and sincere tale} that resonates with truth. \\ 
    \thickhline
      \multicolumn{2}{c}{Movie: ``Lost Boys of Sudan''} \\ 
	\thickhline
      Synopsis & Megan Mylan and Jon Shenk's award-winning documentary Lost Boys of Sudan examines what happens when a pair of Sudanese boys, orphaned due to a civil war in their home country, are allowed to live for a year in the United States. Santito and Peter must contend with extreme examples of culture shock, while also figuring out how to negotiate a world that is physically safe but emotionally and intellectually foreign to them. Unlike many documentaries, the film does not employ a voice-over narration. \\
      Year & 2004 \\
      Genre & Documentary, Special Interest \\
      Director & Megan Mylan, Jon Shenk \\
      Actors & None \\     
      \hline
      {\textsc{\textbf{Gold}}} & 
      The Lost Boys of Sudan works as both a riveting documentary and scathing indictment of colonialism. \\
      {\textsc{\textbf{BertCent}}} & 
      Too short by half, Lost Boys of Sudan affords frustratingly little by way of real analysis and history. But it does introduce us to two extraordinary young men whose faith in this country is as unbearably sad as their stories. \\
      {\textsc{\textbf{BertCent+PtGen}}} & 
      Lost Boys of Sudan is a smart, beautifully filmed \uwave{reflection on love and responsibility}. \\
      {\textbf{\textsc{CondaSum} w/o extracts}} & 
      \emph{\textbf{General}}: An uplifting and enlightening documentary about an earnest street youth. \\
&     \emph{\textbf{Customized (Positive)}}: An {\color{blue!45!blue} uplifting and enlightening documentary} about an earnest street youth. \\
 &    \emph{\textbf{Customized (Negative)}}: It has an {\color{blue!45!blue}intriguing premise and admirable ambitions}, but Lost Boys of Sudan {\color{red!45!red} suffers from an overly maudlin script and a borderline riveting documentary}. \\
      {\textbf{\textsc{CondaSum} w/ extracts}}  & 
      \textit{\textbf{General}}: 
      Lost Boys of Sudan is a powerful and uplifting documentary. \\
     & \textit{\textbf{Customized (Positive)}}: 
     Lost Boys of Sudan is a {\color{blue!45!blue} powerful and uplifting documentary}. \\
 &     \textit{\textbf{Customized (Negative)}}: Lost Boys of Sudan is a {\color{blue!45!blue} powerful and uplifting documentary}. \\ \thickhline
    \end{tabular}%
\end{small}
\caption{Examples of general-purpose and need-specific opinion summaries
  generated by four systems. We also show the consensus summary
  (\textsc{Gold}).  \uwave{Underlined} phrases denote factually
  incorrect information.  Words/phrases in color highlight {\color{blue!45!blue}
    positive} and {\color{red!45!red} negative} aspects.}
  \label{fig:app_ex_sentiment}%
\end{figure*}%

\end{document}